# A Hierarchical Hybrid AI Approach: Integrating Deep Reinforcement Learning and Scripted Agents in Combat Simulations

**Scotty Black, PhD**
**Marine Corps Warfighting Lab**
**Quantico, Virginia**
scotty.black@usmc.mil

**Christian Darken, PhD**
**Naval Postgraduate School**
**Monterey, California**
cjdarken@nps.edu

## ABSTRACT

In the domain of combat simulations in support of wargaming, the development of intelligent agents has predominantly been characterized by rule-based, scripted methodologies with deep reinforcement learning (RL) approaches only recently being introduced. While scripted agents offer predictability and consistency in controlled environments, they fall short in dynamic, complex scenarios due to their inherent inflexibility. Conversely, RL agents excel in adaptability and learning, offering potential improvements in handling unforeseen situations, but suffer from significant challenges such as black-box decision-making processes and scalability issues in larger simulation environments. This paper introduces a novel hierarchical hybrid artificial intelligence (AI) approach that synergizes the reliability and predictability of scripted agents with the dynamic, adaptive learning capabilities of RL. By structuring the AI system hierarchically, the proposed approach aims to utilize scripted agents for routine, tactical-level decisions and RL agents for higher-level, strategic decision-making, thus addressing the limitations of each method while leveraging their individual strengths. This integration is shown to significantly improve overall performance, providing a robust, adaptable, and effective solution for developing and training intelligent agents in complex simulation environments.

## ABOUT THE AUTHORS

**Scotty Black** is a Lieutenant Colonel in the U.S. Marine Corps assigned to the Marine Corps Warfighting Lab (MCWL) as part of the Marine Corps Technical PhD Program. His primary specialty is as an F/A-18 Weapons Systems Officer with over 20 years of Marine Corps experience. LtCol Black is currently the Modeling and Simulation (M&S) and Artificial Intelligence (AI) Technical Advisor for the MCWL Wargaming Division. LtCol Black's experience includes nearly 2,000 flight hours in the F/A-18, graduate of the Weapons and Tactics Instructor (WTI) Course, multiple combat deployments, leading science and technology initiatives for the Marine Corps, a DARPA Service Chiefs Fellowship, and research fellowships at the Naval Information Warfare Center Pacific and the former Space and Naval Warfare Systems Center Pacific.

**Christian Darken** is an Associate Professor in the Department of Computer Science at the Naval Postgraduate School, where he is also a member of the MOVES (Modeling, Virtual Environments and Simulation) faculty. He has more than 35 years of machine learning research experience and has been conducting teaching and research on human behavior models for simulations for over twenty years. His background includes technical program management for Siemens Corporation, and serving as program and general chair for the AAAI-sponsored AIIDE conference.

# A Hierarchical Hybrid AI Approach: Integrating Deep Reinforcement Learning and Scripted Agents in Combat Simulations

**Scotty Black, PhD**

**Marine Corps Warfighting Lab**

**Quantico, Virginia**

scotty.black@usmc.mil

**Christian Darken, PhD**

**Naval Postgraduate School**

**Monterey, California**

cjdarken@nps.edu

## INTRODUCTION

In the domain of combat simulations in support of wargaming, the development of intelligent agents has predominantly been characterized by rule-based, scripted methodologies with deep reinforcement learning (RL) approaches only recently being introduced. Scripted methodologies—a term we use in this paper to generally refer to strategies governed by predefined sets of rules and behaviors—have been instrumental in creating effective, predictable, and logical agents for most environments. However, their rigidity and inability to adapt to unforeseen scenarios or circumstances have typically limited their effectiveness—ultimately leading to predictable outcomes, suboptimal performance, and diminished value when used for wargaming or operational planning.

RL, on the other hand, provides a framework for agents to learn and adapt through direct interactions with their environment, allowing agents to improve their behaviors over time, learn from past experiences, generalize from these experiences, and adapt to changing conditions within the simulation environment. Nevertheless, the application of RL in large combat simulations is not without its challenges, primarily due to the complexity of these environments and the inefficiencies associated with learning in large state spaces. Additionally, the black-box nature of RL models can make the decision-making process opaque, making it difficult to trust and interpret the actions taken by RL agents.

This paper proposes a hierarchical hybrid artificial intelligence (AI) approach that integrates RL and scripted agents in combat simulations to improve performance beyond either approach alone. By structuring the AI system hierarchically, we aim to leverage the strengths of both methods while mitigating their respective weaknesses. Scripted agents are employed to handle well-defined, routine tasks and to provide a consistent baseline behavior at the tactical level, while RL agents are utilized for longer-term decision-making and adaptation in response to evolving circumstances at the operational or strategic level. This hybrid approach aims to create a more robust and effective AI system overall. Specifically, our investigation explores ways to optimize training efficacy given the constraint of limited computational budgets—a typical challenge in the practical application of RL to combat simulations.

## BACKGROUND

### Scripted Agent Decision-Making

Traditional decision-making algorithms such as rule-based systems, behavior trees (BTs), goal-based systems, and finite state machines (FSMs) are examples of approaches central to agent design in games and simulations (Millington, 2006). Most strategy games and combat simulations implementing intelligent agents employ these types of hard-coded methodologies due to their reliability, predictability, and ease of implementation. In this paper, we refer to these types of approaches that do not rely on machine learning as *Scripted Agents*.

Rule-based systems, grounded in if-then logic, offer a methodical framework for agent decision-making. They provide a structured approach, allowing for decisions to be made based on specific, predefined criteria, which results in clear and consistent actions. BTs extend this structure, organizing decisions in a hierarchical manner that mirrors natural decision-making processes, thereby enhancing system flexibility while providing a distinct delineation of decision pathways. This structure not only facilitates easier updates and modifications but also supports varied behavioral patterns. FSMs simplify the representation of agent states, breaking down behaviors into clear, discrete stages with defined transitions, thus facilitating targeted problem-solving. Incorporating goal-based systems into this framework allows agents to pursue specific objectives, allowing them to align their actions and strategies toward achieving these goals. These methodologies together create a strong base for crafting agents that perform reliably and consistently—driven by well-defined rules and logical frameworks—thus allowing for reasonable decision-making across most situations.

However, while this design based on specific domain knowledge can lead to effective and predictable levels of performance in familiar situations, relying on predefined rules, heuristics, and algorithms often comes with inherent inflexibility and rigidity, making them less effective in unexpected or novel situations (Kwasny & Faisal, 1990). The scripted agent's reliance on fixed logic and predetermined pathways often limits the agent's ability to adapt to larger, more dynamic environments (Colledanchise & Ögren, 2018; Millington, 2006)—underscoring the need for more advanced, adaptable AI approaches capable of learning and responding to novel challenges, while still leveraging the consistency and predictability of scripted methodologies.

**Reinforcement Learning**

Reinforcement learning is a subset of machine learning that involves an agent learning to make decisions through direct interaction with its environment. A reinforcement learning problem, shown in Figure 1, typically consists of a decision-maker, referred to as an agent, and an environment, represented by states $s \in S$. The agent takes actions $a_t$ as a function of the current state $s_t$ such that $a_t = A(s_t)$. After choosing an action at time $t(a_t)$, the agent receives a reward $r_{t+1}$ and finds itself in a new state $s_{t+1}$. The action $a_t$ comes from a strategy called a policy $\pi$ that maps states $s \in S$ to a probability of selecting each possible action $\pi(s, a)$. As the agent interacts with the environment, it learns the optimal policy that maximizes its reward in the long run.

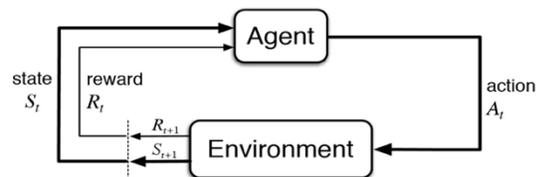

**Figure 1. The Reinforcement Learning Problem**

Notably, RL does not rely on predefined rules or scenarios and instead utilizes a feedback loop where agents learn optimal behaviors through trial and error, guided by rewards or punishments. This learning-based approach enables AI systems to adapt their strategies over time, improving their decision-making capabilities in response to changing conditions and accumulated experience.

RL agents are particularly effective in uncertain environments where the optimal strategy is not known a priori. Nevertheless, while RL has achieved notable successes in games, such as StarCraft (Vinyals et al., 2019), Dota 2 (Berner et al., 2019), Go (Silver et al., 2016), and Atari (Mnih et al., 2015)—achieving expert or even superhuman levels—its application in the more complex realm of combat simulations in support of wargaming has yet to surpass the performance of human or hard-coded agents. Research in RL for combat simulations reveals that, although effective in smaller contexts, scaling up often leads to poor performance (Black & Darken, 2024a, 2024b; Boron & Darken, 2020; Cannon & Goericke, 2020; Coble et al., 2023; Rood, 2022). This issue primarily stems from the exponential increase in state space complexity (Bellman, 1954) and RL's sample inefficiency problem (Mnih et al., 2015), which becomes more pronounced as the complexity of the observation space grows.

**RELATED WORKS**

Existing literature explores the integration of multiple AI methodologies to enhance intelligent agent development, covering areas from RL-enhanced behavior tree development to tactical applications in military simulations and game development. The following review describes current advancements in hybrid AI systems, identifying research gaps and framing our unique contributions to the field.

Ulam et al. (2021) propose combining model-based meta-reasoning with RL to improve the game-playing agent's adaptability in the game of *FreeCiv*. Their approach involves the application of meta-reasoning to guide RL, helping reduce the learning space and learning time by focusing on relevant sub-tasks, leading to improved performance and efficiency as compared to conventional RL agents. While an improvement over more traditional RL approaches, it relies on the need for an accurate and comprehensive meta-reasoning model which may be difficult to construct in games involving complex strategies.

Li et al. (2021) explore the integration of RL and BTs to enhance intelligent agents. Their approach leverages the strengths of both RL and BTs, synchronizing their execution and defining appropriate reward functions to guide effective decision-making. While our research differs in that we do not seek to incorporate RL nodes within our scripted agents' BTs, their research offers valuable insights into integrating RL with rule-based agents, particularly in managing complex and hierarchical task assignments.

Zhao et al. (2023) propose a method for generating interpretable BTs from RL policies, aiming to enhance the interpretability of RL in applications such as robotics and gaming. This involves translating RL policies directly into BTs, addressing the challenge of making RL decisions understandable and manageable. While this approach is indeed valuable in informing our research, our hybrid approach involves the incorporation of RL and scripted agents during gameplay, rather than relying on RL to generate a scripted agent.

Yu et al. (2023) present an enhanced decision-making method for combat simulations through the integration of RL and supervised learning by employing an improved Multi-Agent Deep Deterministic Policy Gradient (MADDPG) framework. It showcases significant performance improvements, underlining the benefits of combining RL with additional AI techniques to help address large action spaces and sparse rewards. While this is an improved approach to RL, it differs from our approach in that we use a hybrid approach both in the training and control of our agents, rather than just in the training of our agents.

Further, research by Hu et al. (2021), Bignold et al. (2021), and Kallstrom and Heintz (2020) navigate the tactical applications of RL in military simulations, demonstrating the practical implications for air combat maneuver planning and persistent rule-based interactions. On the game development front, Zhao et al. (2020), Noblega et al. (2019), and Kim and Ahn (2018) explore AI's role in game balancing and development, while Vazquez-Nunez et al. (2020), Joppen et al. (2018), and Ponsen (2006) introduce hybrid computational intelligence, showcasing the efficacy of combining various methodologies for adaptive game tactics.

While the foundational insights from the literature offer a broad spectrum of approaches to advancing AI in domains relevant to combat simulations, our research distinguishes itself by advancing a hierarchical hybrid model that combines the strengths of RL with the precision of scripted systems. This approach not only addresses challenges of adaptability and decision-making in complex environments but also proposes a unique solution to the integration of two different AI systems, aiming to enhance both the strategic flexibility and tactical execution of AI agents in combat simulations.

## METHODOLOGY

To assess the efficacy of using a hierarchical hybrid AI approach that integrates an RL manager agent with scripted subordinate agents, we employ the following methodology.

### Atlatl Combat Simulation Environment

We use the Atlatl Simulation Environment (Darken, 2022) to develop, implement, and experiment with our research approach. Atlatl is a simple but effective combat simulation developed at the Naval Postgraduate School. It includes an underlying combat model that is purposefully simplistic and built to facilitate rapid AI experimentation. In particular, the Gymnasium (Farama Foundation, 2023) API is implemented supporting interfacing with standard RL codebase and algorithms, such as Stable Baselines 3 (2024). This type of basic environment allows researchers to develop, apply, and evaluate cutting-edge AI to operational and tactical problems more efficiently and effectively than using operational or high-fidelity simulation systems.

Through a web browser interface, Atlatl allows a human player to play against the AI; however, the simulation can also run headless with an AI playing against another AI. Additionally, a browser-based replay capability allows for replays of AI versus AI engagements. An example of a simple scenario on an Atlatl gameboard is shown in Figure 2. Units are represented visually by their respective military operational terms and graphics, and terrain is represented visually in colors (e.g., water is blue, rough terrain is brown, urban is gray).

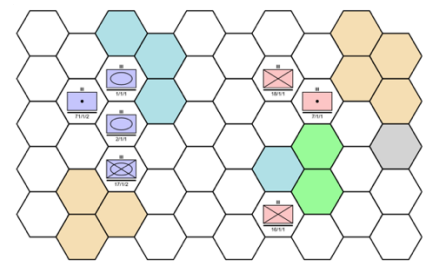

**Figure 2. Atlatl Gameboard Example**

The scoring system within Atlatl is fully customizable and is given from the perspective of the blue player. For this experiment, scoring—which is our performance metric—is computed based on kills, losses, and control of urban areas (used to represent cities). We use the following scoring function:

$$S_{total} = S_{b\_city} + S_{b\_combat} - (S_{r\_city} + S_{r\_combat})$$

Where $S_{b\_city}$ is the score per city owned by the blue faction; $S_{b\_combat}$ is the score for each red unit damaged in combat by the blue faction; $S_{r\_city}$ is the score per city owned by the red faction, and $S_{r\_combat}$ is the score for each blue unit damaged in combat by the red faction.

Control of urban hexagons plays a significant role in calculating the player's score. Initially, these hexagons are neutral, without any faction's control. A faction gains control and starts accumulating points by positioning a unit on an urban hexagon. Specifically, if the scenario contains a single urban hexagon, control is valued at 24 points per phase, where there are two phases (blue action and red action) per complete turn. In scenarios with two urban hexagons, each is worth 12 points per phase. Of note, once an urban hexagon is occupied, it remains under that faction's control even if the unit vacates the hexagon, up until a unit of the opposing faction occupies the same urban hexagon.

Regarding combat, each unit begins with an initial 100 strength points. Each damage point inflicted on a red unit translates into a positive point for the blue faction, while each damage point inflicted on a blue unit translates into a negative point for the blue faction. If a unit's strength drops below 50 points, it is removed from the game (i.e., deemed ineffective) and the remaining strength points are awarded to the opposing faction.

**Scripted Agent**

In our experiment, the *Scripted Agent* uses a behavior tree where it first assesses its posture as either "offensive" or "defensive" based on the relative strength of its faction as compared to its opponent. For each movable unit, it first checks if there is an enemy unit within its attack range; if so, it uses a uniform distribution to select a unit to attack. If no targets are available, the unit calculates movement based on a scoring system, which is influenced by its posture. It then evaluates each possible hexagon it can move to based on its proximity to enemy units and urban hexagons. Finally, the unit selects the move leading to the hexagon with the lowest score as it seeks to minimize the distance to its objective.

This system allows for rapid, deterministic responses to changing game dynamics, adhering to a set of predefined rules and scripts. Furthermore, the *Scripted Agent* exhibits goal-oriented behavior, albeit in a basic form, aiming to achieve specific objectives such as capturing key terrain or defeating enemy units. While not leveraging complex planning algorithms, this goal-oriented approach ensures that the agent's actions are consistently directed toward achieving its overarching strategic objectives of maximizing the game score. While a simple model, it has proven to be an effective agent that takes human-like actions and regularly achieves reasonably good scores.

**Integration with Reinforcement Learning**

To address the limitations of both scripted and RL agents, our approach introduces an *RL Manager Agent* that operates in conjunction with scripted units, which we will refer to as *Scripted Subordinate Agents*. We designate this integrated system, consisting of an *RL Manager Agent* and its assigned *Scripted Subordinate Agents,* as a *Hybrid Agent*. This *Hybrid Agent* allows for the combination of strengths of both approaches: the reliability and known quantity of rule-based decision-making at the tactical level, and the adaptability and optimization capabilities of RL at the strategic level. Of note, while the *Scripted Subordinate Agents* are actual entities on the gameboard, the *RL Manager Agent* is not an actual entity on the gameboard, but rather an abstract agent that issues objective areas to guide its *Scripted Subordinate Agents*.

When using the *Hybrid Agent* model, at scenario initiation, after units are randomly created per a given scenario seed, each unit is assigned to its respective *RL Manager Agent*. Each *RL Manager Agent* is initially assigned and controls 3 units. For example, if the Blue faction starts with 6 units, a total of 2 individual *RL Manager Agents* are created to control 3 units each. During gameplay, the *RL Manager Agent* operates by selecting an objective area and passing this to each of its *Scripted Subordinate Agents*. Each *RL Manager Agent's* objective area is essentially a super "hexagon" in that it contains a center hexagon and two surrounding layers of hexagons. An example is shown in Figure 3, where the *RL Manager's Objective Area* is denoted by the blue area.

Our implementation of *RL Manager Agents* is derived from that of Sutton et al.'s (1999) "options" framework. "Options" are a form of temporal abstraction that represent a sequence of actions aimed at achieving a specific sub-goal within a larger problem, allowing for more complex, hierarchical decision-making processes.

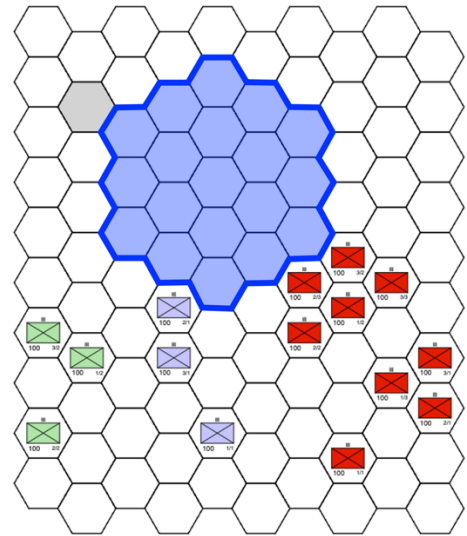

**Figure 3. RL Manager's Objective Area**

Unlike the *Scripted Subordinate Agents*, *RL Manager Agents* do not make decisions at each time step in the game. Instead, the *RL Manager Agent's* temporal abstraction is variable and event-driven. Each *RL Manager Agent* is only called to make a decision when all of its units (that are not ineffective) arrive within its objective area. At this point, the respective *RL Manager Agent* generates its own objective area independent of the other *RL Manager Agents* that may also exist. This mirrors the "options" framework (Sutton et al., 1999), where decisions span over multiple time steps, focusing on achieving broader objectives, rather than responding to every single environmental change.

Conversely, *Scripted Subordinate Agents* make decisions at each time step using one of two behavior modules. First, the *Scripted Subordinate Agent* checks to see if it is within its assigned objective area. If the *Scripted Subordinate Agent* is not within its objective area, it activates its *Move Modu*le, where it simply selects the hexagon to move to that is closest to its objective area. However, if an attack opportunity exists (i.e., an adversary unit is within its attack range), it will attack the opposing faction's unit instead of moving. Once the *Scripted Subordinate Agent* detects that it is within its objective area, it activates its *Fight module,* where the unit executes the scripted behavior algorithm described above. Of note, we restrict the information available to the *Scripted Subordinate Agent* to that of its objective area. In other words, the game's global state space is culled to only include the information within the *Scripted Subordinate Agent's* objective area.

**Gymnasium Environment**

For our RL training, we use a custom Gymnasium (Farama Foundation, 2023) environment configurable for different roles (blue or red), AI types, and scenarios. The action space of our *RL Manager Agent* is defined as the set of super hexagons whose center hexagons contain the centers of a uniform $7 \times 7$ square grid superimposed over the scenario map, resulting in a total of 49 discrete actions.

**Neural Network Architecture**

We use a residual convolutional neural network (CNN) designed to process an input observation space of any size. The architecture uses convolutional layers to transform this input observation into 64 output channels. This is followed by 7 additional layers of 64 channels each. Each layer features pointwise convolutions with a kernel size of $1 \times 1$ and a stride of 1. Additionally, in each layer, we include a Rectified Linear Unit (ReLU) activation function and a residual connection. After 7 layers, the resulting multi-dimensional tensor is then flattened into a one-dimensional tensor and is passed through a final linear layer. This layer maps the flattened tensor to a 512-dimensional feature vector, which is then passed through a final ReLU activation function.

**Reinforcement Learning Algorithm**

We employ the Deep Q-Network (DQN) algorithm (Raffin, 2018). The hyperparameters used were optimized through extensive hyperparameter tuning in similar scenarios, though not specific to this experiment. The final configuration includes a learning rate of 0.0002, a buffer size of 1,000,000, learning starting at 10,000 steps, a batch size of 64, and a discount factor ($\gamma$) of 0.93. The target network update interval is set to 1,000 steps. For exploration, we employ an initial epsilon ($\epsilon_i$) of 1.0, decaying linearly to a final epsilon ($\epsilon_f$) of 0.01, with an exploration fraction of 1.0. The training frequency is set to every 4 steps, with 1 gradient step taken per training update.

**Manager's Observation Abstraction**

The default RL agent's global observation in Atlatl consists of an $18 \times n \times m$ tensor, where $n$ and $m$ are the height and width of the gameboard. Each channel of the tensor represents one specific type of information to be captured, which is generally, but not always, represented as a binary matrix.

Rather than providing the *RL Manager Agent* the entire observation space, we modify a few of the default feature extractors as well as abstract the resulting observation in a map size-invariant way to provide the *RL Manager Agent* with a more tractable, compressed observation space. We do this by taking the full three-dimensional observation tensor and applying a coarse abstraction, which results in a final observation space of $17 \times 7 \times 7$. This reduction involves reducing the full tensor into a smaller grid by comparing the original and reduced grid sizes to ensure proportional representation. Each cell in the original grid is then mapped to the reduced grid based on the overlap, with values adjusted to reflect the proportionate overlapped area, preserving the original spatial information. This method provides the RL Manager Agent with a simplified yet accurate state space representation, enhancing its processing efficiency without significant loss of detail.

As shown in Figure 4, channel 0 depicts the blue units that belong to the current *RL Manager Agent* on-move; channel 1 depicts the other *RL Manager Agents'* objective areas; channels 2 and 3 are matrices that depict the health level of each unit on the gameboard based on factions; channels 4 through 7 depict unit types; channels 8 through 12 depict terrain types; channels 13 and 14 depict the respective city owners (i.e., which faction was the last to pass through an urban hexagon); channel 15 is a matrix filled with a phase indicator value representing the current phase of the game; and channel 16 is a matrix filled with the normalized game score. While we recognize that these last two features can be represented more compactly as vectors or scalars rather than matrices, we maintain the matrix construct in this study for simplicity.

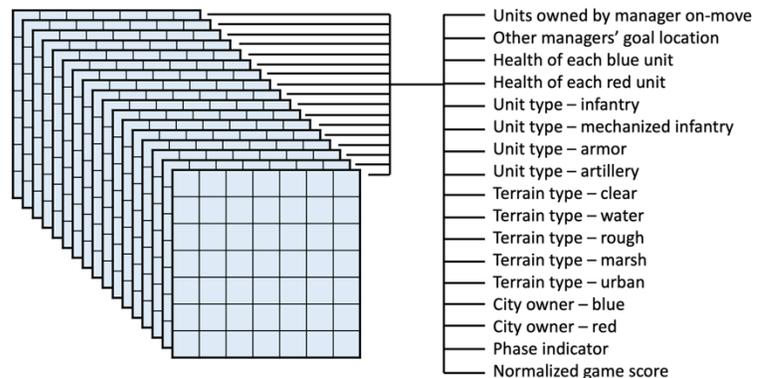

**Figure 4. RL Manager's Observation**

**Scenarios**

We use randomly generated scenarios consisting of $10 \times 10$ hexagonal gameboards. Each game initiates with a randomly assigned number of units per faction, specifically either 6 or 9 units per faction. This results in either 2 or 3 managers per faction, respectively. Each scenario also includes either 1 or 2 urban hexagons randomly placed according to force ratio. If one faction has a smaller force ratio (i.e., fewer units as compared to the opposing faction), the urban hexagon(s) is placed on its side of the gameboard. If the force ratios are equal (i.e., both factions have an equal number of units), the urban hexagon(s) is placed in a neutral location along the middle axis of the board. The number of phases in the game is set to 40, where each phase is one entire turn for one faction (i.e., one faction is allowed to make one legal move for each of its available entities). Setting the number of phases to this value provides enough turns for a unit to go from one end of the gameboard to the opposite end and return, likely giving them enough turns to execute complex maneuvering if warranted. Two examples of scenario initial starting conditions are shown in Figure 5. The Blue faction is color-coded blue, green, and yellow to indicate they belong to separate Managers. The red faction is color-coded red. Urban hexagons are color-coded grey.

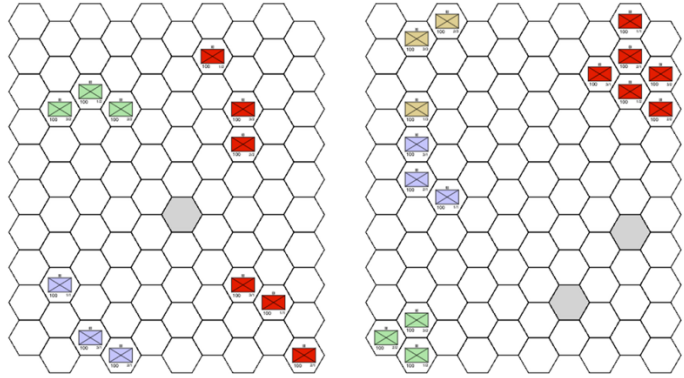

**Figure 5. Example Scenarios**

### Reinforcement Learning Training

Our training is designed to test the robustness and generalizability of the *Hybrid Agent* in a controlled yet varied environment. We use randomly generated scenarios employing a scenario cycle of 10, where the simulation generates 10 random scenarios and then continuously cycles through these 10 scenarios. This approach introduces a level of variability and unpredictability into the training process without overwhelming the *RL Manager Agent* with infinite possibilities, ensuring that the agent encounters a variety of situations but also gets the chance to learn from repeated exposure to the same scenarios. We believe that a cycle of 10 provides a balance between novelty and repetition, aiding in the deeper learning of strategies that are effective across different scenarios while also adapting to specific challenges presented within each scenario.

We use a training budget of 10 million steps for our *RL Manager Agent* and train this model against the baseline scripted model described previously as the adversary agent. By training against a baseline scripted adversary model, we can expose the resulting *Hybrid Agent* to a consistent set of strategies and behaviors, which helps in understanding how well this approach can adapt and counteract a known set of tactics.

Additionally, to ensure that our results are not overly dependent on a particular set of starting conditions, we use a different random seed to train 5 different *RL Manager Agents*. Different seeds result in a different set of scenarios and variations in learning experiences, which, when aggregated, offer a more comprehensive understanding of the model's performance.

To learn effective behaviors, we design a reward system that balances defeating the opposing faction and occupying urban hexagons with preserving its own force. This reward is computed each time a manager is called to make a decision using the following equation.

$$R_{engineered} = \max(R_m - P_g, 0)\frac{S_{m,c}}{S_{m,o}} + B_t I_t$$

Where $m$ represents a specific manager; $R_m$ is the accumulated manager $m$'s score based on its units' kills, losses, and urban hexagons held; $P_g$ is a penalty of 25 points for assigning the same objective area as one or more of the other managers (which discourages managers from assigning the same objective area unless it is overwhelmingly advantageous to do so); $S_{m,c}$ is the current total unit strength for manager $m$; $S_{m,o}$ is the original total unit strength for manager $m$; $B_t$ is a terminal bonus reward of 25 points that our research shows discourages units from moving into the adversary units' attack range during the last turn of the game; and $I_t$ is a terminal game state indicator that takes on a value of 1 if the game is terminal or 0 if the game is not terminal.

The learning curves for each of the 5 seeds used are shown in Figure 6. The graph on the left depicts *Mean Rewards vs. Training Steps* for the *RL Individual Agents*, and the graph on the right is for the *RL Manager Agents*. The graph shows that all agents have yet to plateau when reaching our training budget of 10 million training steps. Of note, the *RL Individual Agents* are evaluated every 100,000 training steps while the *RL Manager Agents* are evaluated every 10,000 training steps. This does not affect learning, but it is the reason for the reduced number of datapoints on the left graph as compared to the right graph. As depicted, all of the *RL Manager Agents* still appear to be learning at training termination, indicating that if we afforded these agents a larger training budget, performance would likely continue to improve.

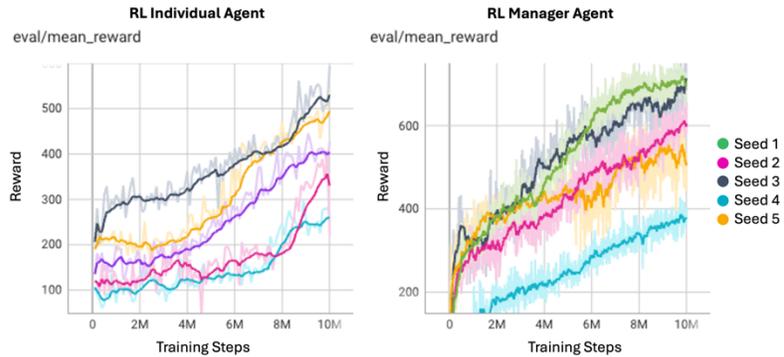

**Figure 6. RL Agent Training Learning**

We also train *RL Individual Agents* to include as part of our analysis. Unlike the *RL Manager Agent*, which controls groups of units via objective areas, *RL Individual Agents* control the game actions of individual units (i.e., instead of a script delineating the unit's action, the RL model selects the action). Including this model as part of our evaluation allows us to compare our hybrid approach to the two models we integrate and look to improve upon. Of note, previous experiments we have conducted have shown success when training R*L Individual Agents* using smaller gameboard sizes of up to $5 \times 5$. However, with RL, increasing complexity in the state space typically results in a nonlinear increase in training steps needed to achieve the same level of performance. Thus, we do not anticipate achieving good results from the *RL Individual Agents*.

The development and training of these *RL Individual Agents* is similar to that of our *RL Manager Agents* with three key differences. The first key difference is that the *RL Individual Agent* uses an observation space of $18 \times n \times m$ where $n \times m$ are the dimensions of the gameboard, which in our case is a $10 \times 10$. Channels 2 through 16 remain the same as with the *RL Individual Agent* observation but with no coarse abstraction applied. In addition to these 15 channels, the *RL Individual Agents* has a channel that depicts where the blue unit to be moved (or on-turn) is located; a channel depicting all blue units that have the ability to move during the current phase; and a channel depicting all legal moves available to the unit on-move.

The second key difference is that while we use the same neural network architecture as with the *RL Manager Agent*, we use HexagDLy (Steppa & Holsch, 2019) hexagonal convolutions rather than standard convolutions in each layer.

The last key difference is the engineered reward equation:

$$R_{engineered} = \max(R_m - P_g, 0)\frac{S_c}{S_o} + B_t I_t$$

Where $R_{raw}$ is the difference in game score between the current time step and the previous time step. The rest of the variables are as described in the Manager's engineered reward equation except not tied to a specific manager but instead based on the entire faction.

**Model Evaluation**

We evaluate each of our trained *RL Manager Agent* models using the scripted model described previously as the adversary. We run 100,000 games using the same respective scenario seed that was used for training, recording each score. We conduct three distinct evaluations for each scenario seed used: *Hybrid Agent* vs. *Scripted Agent*; *RL Individual Agent* vs. *Scripted Agent*; and *Scripted Agent* vs. *Scripted Agent*. The performance of the *Scripted Agent* and the *RL Individual Agent* serve as our benchmarks to assess if our *Hybrid Agent* approach is more effective than either or both of these more traditional approaches.

## RESULTS AND DISCUSSION

The results of our evaluation are summarized in Table 1 and illustrated in Figure 7. Table 1 depicts the mean scores $\pm$ the standard error of the mean (SEM) across the 100,000 evaluation games for each random seed used. Figure 7 shows the boxplots for each model across the different seeds.

**Table 1. Model Performance Results**

| | Mean Scores $\pm$ SEM | | |
|---|---|---|---|
| Random Seed | Scripted | RL Individual | Hybrid |
| 1 | $-19.9 \pm 2.5$ | $-572.3 \pm 1.2$ | $692.1 \pm 3.0$ |
| 2 | $-63.7 \pm 2.7$ | $-366.2 \pm 2.2$ | $671.1 \pm 2.4$ |
| 3 | $162.1 \pm 3.1$ | $-334.0 \pm 2.3$ | $580.4 \pm 3.4$ |
| 4 | $-117.2 \pm 2.1$ | $-693.2 \pm 0.0$ | $13.3 \pm 2.8$ |
| 5 | $262.9 \pm 2.5$ | $-1051.1 \pm 0.9$ | $493.1 \pm 3.3$ |
| Overall Mean | $-0.904$ | $-603.371$ | $489.998$ |

To confirm statistical significance, we conduct a Paired-Sample T-Test, setting our $\alpha$ to 0.05, and comparing the mean scores between *Hybrid Agent* vs. *Scripted Agent*; *Scripted Agent* vs. *Scripted Agent*; and *RL Individual Agent* vs. *Scripted Agent* across each of the 5 random seeds, for a total of 15 pairwise comparisons. Each of these returned a p-value of $< 0.001$, indicating statistical significance across all comparisons.

Our analysis shows that the *Hybrid Agent* outperforms both the *Scripted Agent* and the *RL Individual Agent* by large margins across the board. Across all five seeds, the *Hybrid Agent* finished with a mean score of 489.998, while the *Scripted Agent* baseline hovered near zero ($-0.904$) and the *RL Individual Agent* collapsed to $-603.371$. In practical terms, the hybrid approach led to more red strength destroyed, fewer blue units lost, and urban hexes captured earlier and for longer. Visual replays of all scenarios also confirmed superior performance of the *Hybrid Agent*.

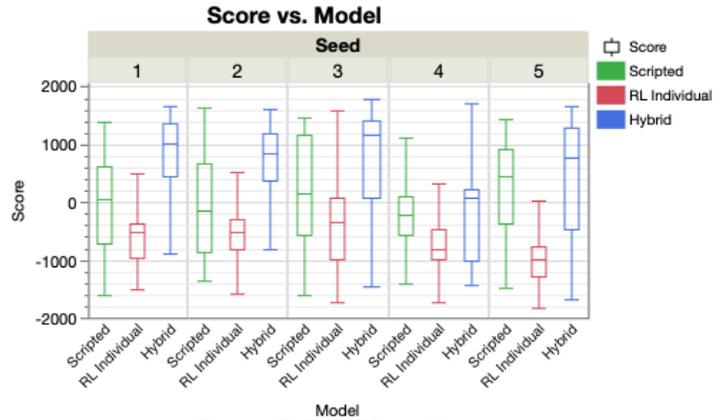

**Figure 7. Box Plot of Scores**

These results demonstrate the *Hybrid Agent's* ability to learn and apply strategies across a number of diverse scenarios. The learning curves of the *RL Manager Agents* show a consistent improvement in performance, suggesting that with extended training or increased computational resources, better performance is still achievable.

Furthermore, our analysis suggests that by delegating specific tactical decisions to scripted agents, the RL component is freed to concentrate on higher-level strategic considerations, thus reducing the computational burden and enhancing overall system performance. The RL manager sees a greatly compressed state-action space and can focus its learning budget on higher-value strategic choices (where to fight, when to mass forces, when to avoid contact). This potentially reduces sample complexity and accelerates convergence, which not only streamlines the decision-making process but also improves the scalability of the model to handle more complex and larger-scale scenarios. Additionally, because the manager's *objective area* option spans many game time ticks, rewards arrive only when all assigned units converge on that area, which creates a natural credit-assignment window and discourages thrashing.

These results also show that the integration of RL with scripted agents can lead to a more nuanced and flexible response mechanism, allowing the *Hybrid Agent* to effectively adapt to changing conditions and unforeseen challenges more quickly by allowing developers to independently alter either the scripted behavior algorithms or the RL training mechanisms. This blend of predictability and adaptability is particularly valuable in the context of military simulations, where both qualities are essential for realistic and effective strategy formulation.

**CONCLUSION**

This study demonstrates that employing a hierarchical hybrid AI approach—RL for the "where and when" and scripts for the "how"—our model can outperform both traditional RL and scripted strategies when applied independently in combat simulations. Of note, this approach does not remove the need for expert-crafted behaviors; rather, it elevates them by letting a machine learning-driven manager decide which how to best employ its units.

For our implementation, we find that the integration of RL with scripted agents effectively mitigates the inherent limitations of each method when used separately. By employing a hierarchical structure, our model utilizes RL for higher-level strategic decision-making while employing scripted agents to manage routine, lower-level tactical decisions. This two-level approach not only improves overall performance but also offers a scalable solution capable of adapting to increasingly complex scenarios.

Nevertheless, our research is not without its challenges. We acknowledge existing hurdles, particularly concerning computational efficiency and the scalability of RL algorithms in general and within our model. Addressing these issues should be a primary focus for future work. We plan to refine and extend our framework to include a wider array of scenarios, incorporating broader and diverse terrain and unit types, thereby closer approximating more realistic applications.

The insights gained from this study offer a promising path forward for the development of AI-driven agents in operational-level combat simulations, real-time strategy games, and other similar domains. By combining the predictable precision of scripted systems with the dynamic adaptability of RL, our model advances the research of intelligent agents in complex simulation environments and embodies a novel integration of established and emerging AI methodologies—promising to elevate the capabilities and performance of intelligent systems in this challenging domain.

**REFERENCES**

Bellman, R. (1954). The Theory of Dynamic Programming. *Bulletin of the American Mathematical Society*, *60*, 503–515. https://apps.dtic.mil/sti/citations/AD0604386

Berner, C., Brockman, G., Chan, B., Cheung, V., Dennison, C., Farhi, D., Fischer, Q., Hashme, S., Hesse, C., Józefowicz, R., Gray, S., Olsson, C., Pachocki, J., Petrov, M., Salimans, T., Schlatter, J., Schneider, J., Sidor, S., Sutskever, I., … Zhang, S. (2019). *Dota 2 with Large Scale Deep Reinforcement Learning*. https://arxiv.org/abs/1912.06680

Bignold, A., Cruz, F., Dazeley, R., Vamplew, P., & Foale, C. (2021). *Persistent Rule-based Interactive Reinforcement Learning* (arXiv:2102.02441). arXiv. http://arxiv.org/abs/2102.02441

Black, S., & Darken, C. (2024a). Localized Observation Abstraction Using Piecewise Linear Spatial Decay for Reinforcement Learning in Combat Simulations. *MODSIM WORLD 2024 Conference Publications*. MODSIM WORLD 2024, Norfolk, VA. http://modsimworld.org/papers/2024/MODSIM_2024_paper_12.pdf

Black, S., & Darken, C. J. (2024b). A multi-model approach to modeling intelligent combat behaviors for wargaming. In P. J. Schwartz, M. E. Hohil, & B. Jensen (Eds.), *Artificial Intelligence and Machine Learning for Multi-Domain Operations Applications VI* (p. 28). SPIE. https://doi.org/10.1117/12.3015088

Boron, J., & Darken, C. (2020). Developing Combat Behavior through Reinforcement Learning in Wargames and Simulations. *2020 IEEE Conference on Games (CoG)*, 728–731. https://doi.org/10.1109/CoG47356.2020.9231609

Cannon, C. T., & Goericke, S. (2020). *Using Convolution Neural Networks to Develop Robust Combat Behaviors Through Reinforcement Learning* [Naval Postgraduate School]. https://apps.dtic.mil/sti/trecms/pdf/AD1150887.pdf

Coble, J., Barton, A., Darken, C., & Black, S. (2023). Optimizing Naval Movement Using Deep Reinforcement Learning. *2023 22nd IEEE International Conference on Machine Learning and Applications (ICMLA)*. 2023 International Conference on Machine Learning and Applications (ICMLA), Jacksonville, FL, USA. https://doi.org/10.1109/ICMLA58977.2023.00062

Colledanchise, M., & Ögren, P. (2018). *Behavior Trees in Robotics and AI: An Introduction*. https://doi.org/10.1201/9780429489105

Darken, C. (2022). *Atlatl* [Computer software]. https://gitlab.nps.edu/cjdarken/atlatl

Farama Foundation. (2023). *Gymnasium Documentation*. Gymnasium Documentation. https://gymnasium.farama.org

Hu, D., Yang, R., Zuo, J., Zhang, Z., Wu, J., & Wang, Y. (2021). Application of Deep Reinforcement Learning in Maneuver Planning of Beyond-Visual-Range Air Combat. *IEEE Access*, *9*, 32282–32297. https://doi.org/10.1109/ACCESS.2021.3060426

Kallstrom, J., & Heintz, F. (2020). Agent Coordination in Air Combat Simulation using Multi-Agent Deep Reinforcement Learning. *2020 IEEE International Conference on Systems, Man, and Cybernetics (SMC)*, 2157–2164. https://doi.org/10.1109/SMC42975.2020.9283492

Kim, M.-J., & Ahn, C. W. (2018). Hybrid Fighting Game AI Using a Genetic Algorithm and Monte Carlo Tree Search. *Proceedings of the Genetic and Evolutionary Computation Conference Companion*, 129–130. https://doi.org/10.1145/3205651.3205695

Kwasny, S. C., & Faisal, K. A. (1990). Overcoming Limitations of Rule-Based Systems: An Example of a Hybrid Deterministic Parser. In G. Dorffner (Ed.), *Konnektionismus in Artificial Intelligence und Kognitionsforschung* (Vol. 252, pp. 48–57). Springer Berlin Heidelberg. https://doi.org/10.1007/978-3-642-76070-9_5

Li, L., Wang, L., Li, Y., & Sheng, J. (2021). Mixed Deep Reinforcement Learning-behavior Tree for Intelligent Agents Design: *Proceedings of the 13th International Conference on Agents and Artificial Intelligence*, 113–124. https://doi.org/10.5220/0010316901130124

Millington, I. (2006). *Artificial Intelligence for Games*. Elsevier.

Mnih, V., Kavukcuoglu, K., Silver, D., Rusu, A. A., Veness, J., Bellemare, M. G., Graves, A., Riedmiller, M., Fidjeland, A. K., Ostrovski, G., Petersen, S., Beattie, C., Sadik, A., Antonoglou, I., King, H., Kumaran, D., Wierstra, D., Legg, S., & Hassabis, D. (2015). Human-Level Control Through Deep Reinforcement Learning. *Nature*, *518*(7540), 529–533. https://doi.org/10.1038/nature14236

Noblega, A., Paes, A., & Clua, E. (2019). Towards Adaptive Deep Reinforcement Game Balancing: *Proceedings of the 11th International Conference on Agents and Artificial Intelligence*, 693–700. https://doi.org/10.5220/0007395406930700

Ponsen, M. J. V. (2006). Automatically Generating Game Tactics through Evolutionary Learning. *AI Magazine*, *27*(3). https://doi.org/10.1609/aimag.v27i3.1894

Rood, P. R. (2022). *Scaling Reinforcement Learning Through Feudal Muti-Agent Hierarchy* [Naval Postgraduate School]. https://apps.dtic.mil/sti/trecms/pdf/AD1201723.pdf

Silver, D., Huang, A., Maddison, C. J., Guez, A., Sifre, L., van den Driessche, G., Schrittwieser, J., Antonoglou, I., Panneershelvam, V., Lanctot, M., Dieleman, S., Grewe, D., Nham, J., Kalchbrenner, N., Sutskever, I., Lillicrap, T., Leach, M., Kavukcuoglu, K., Graepel, T., & Hassabis, D. (2016). Mastering the Game of Go with Deep Neural Networks and Tree Search. *Nature*, *529*(7587), 484–489. https://doi.org/10.1038/nature16961

Stable Baselines3. (2024). *Stable-Baselines3 Docs—Reliable Reinforcement Learning Implementations*. https://stable-baselines3.readthedocs.io/en/master/index.html

Steppa, C., & Holsch, T. L. (2019). HexagDLy—Processing Hexagonally Sampled Data with CNNs in PyTorch. *SoftwareX*, *9*(2352–7110), 193–198. https://doi.org/10.1016/j.softx.2019.02.010

Sutton, R. S., Precup, D., & Singh, S. (1999). Between MDPs and Semi-MDPs: A Framework for Temporal Abstraction in Reinforcement Learning. *Artificial Intelligence*, *112*(1–2), 181–211. https://doi.org/10.1016/S0004-3702(99)00052-1

Ulam, P., Jones, J., & Goel, A. (2021). Combining Model-Based Meta-Reasoning and Reinforcement Learning For Adapting Game-Playing Agents. *Proceedings of the AAAI Conference on Artificial Intelligence and Interactive Digital Entertainment*, *4*(1), 132–137. https://doi.org/10.1609/aiide.v4i1.18685

Vazquez-Nunez, A. E., Fernandez-Leiva, A. J., Garcoa-Sanchez, P., & Mora, A. M. (2020). Testing Hybrid Computational Intelligence Algorithms for General Game Playing. In P. A. Castillo, J. L. Jimenez Laredo, & F. Fernandez De Vega (Eds.), *Applications of Evolutionary Computation* (Vol. 12104, pp. 446–460). Springer International Publishing. https://doi.org/10.1007/978-3-030-43722-0_29

Vinyals, O., Babuschkin, I., Czarnecki, W. M., Mathieu, M., Dudzik, A., Chung, J., Choi, D. H., Powell, R., Ewalds, T., Georgiev, P., Oh, J., Horgan, D., Kroiss, M., Danihelka, I., Huang, A., Sifre, L., Cai, T., Agapiou, J. P., Jaderberg, M., … Silver, D. (2019). Grandmaster Level in StarCraft II Using Multi-Agent Reinforcement Learning. *Nature*, *575*(7782), 350–354. https://doi.org/10.1038/s41586-019-1724-z

Yu, S., Zhu, W., & Wang, Y. (2023). Research on Wargame Decision-Making Method Based on Multi-Agent Deep Deterministic Policy Gradient. *Applied Sciences*, *13*(7), 4569. https://doi.org/10.3390/app13074569

Zhao, C., Deng, C., Liu, Z., Zhang, J., Wu, Y., Wang, Y., & Yi, X. (2023). Interpretable Reinforcement Learning of Behavior Trees. *Proceedings of the 2023 15th International Conference on Machine Learning and Computing*, 492–499. https://doi.org/10.1145/3587716.3587798

Zhao, Y., Borovikov, I., De Mesentier Silva, F., Beirami, A., Rupert, J., Somers, C., Harder, J., Kolen, J., Pinto, J., Pourabolghasem, R., Pestrak, J., Chaput, H., Sardari, M., Lin, L., Narravula, S., Aghdaie, N., & Zaman, K. (2020). Winning is Not Everything: Enhancing Game Development with Intelligent Agents. *IEEE Transactions on Games*, *12*(2), 199–212. https://doi.org/10.1109/TG.2020.2990865